\documentclass[conference]{IEEEtran}
\IEEEoverridecommandlockouts
\usepackage{cite}
\usepackage{amsmath,amssymb,amsfonts}
\usepackage{graphicx}
\usepackage{textcomp}
\usepackage{xcolor}
\def\BibTeX{{\rm B\kern-.05em{\sc i\kern-.025em b}\kern-.08em
    T\kern-.1667em\lower.7ex\hbox{E}\kern-.125emX}}

\usepackage{amssymb}
\usepackage{amsmath}
\usepackage{commath}
\usepackage{mathtools}
\usepackage{subcaption}
\usepackage[linesnumbered,ruled]{algorithm2e}
\usepackage{algpseudocode}
\usepackage{array}
\usepackage{booktabs}
\begin{document}

\title{Joint Debiased Representation Learning and \\ Imbalanced Data Clustering}

\author{Mina Rezaei, Emilio Dorigatti, David Ruegamer, Bernd Bischl \\
\{mina.rezaei,emilio.dorigatti,david.ruegamer,bernd.bischl\}@stat.uni-muenchen.de
\\
Statistical Learning and Data Science, Ludwig-Maximilians-University Munich } 

\maketitle

%----------------------- ABSTRACT-----------------%
\begin{abstract} \label{abstract}
One of the most promising approaches for unsupervised learning is combining deep representation learning and deep clustering. Some recent works propose to simultaneously learn representation using deep neural networks and perform clustering by defining a clustering loss on top of embedded features. However, these approaches are sensitive to imbalanced data and out-of-distribution samples. As a consequence, these methods optimize clustering by pushing data close to randomly initialized cluster centers. This is problematic when the number of instances varies largely in different classes or a cluster with few samples has less chance to be assigned a good centroid. To overcome these limitations, we introduce a new unsupervised framework for joint debiased representation learning and image clustering. We simultaneously train two deep learning models, a deep representation network that captures the data distribution, and a deep clustering network that learns embedded features and performs clustering. Speciﬁcally, the clustering network and learning representation network both take advantage of our proposed statistics pooling block that represents mean, variance, and cardinality to handle the out-of-distribution samples and class imbalance. Our experiments show that using these representations, one can considerably improve results on imbalanced image clustering across a variety of image datasets. Moreover, the learned representations generalize well when transferred to the out-of-distribution dataset. 

\end{abstract}


%----------------------- INTRODUCTION-----------------%
\section{Introduction} \label{introduction}

%Motivation for Unsupervised
Learning from unlabelled data can reduce development costs in many deep learning applications that otherwise require annotations from experts such as medical image diagnosis and autonomous driving. Clustering is one of the most fundamental methods in unsupervised learning, grouping observations by similar features without supervision or prior knowledge of the nature of the clusters. Unsupervised cluster algorithms have been investigated extensively over the last years in terms of underlying distance functions, feature selection, and different grouping algorithms.

%Definition of between and within cluster imbalanced 
Some existing clustering approaches implicitly assume that the clusters share certain properties, at least within certain boundaries~\cite{raykov2016k}. For example, clusters are assumed to only diverge within a certain range, or, the scale of the cluster spread is presumed to be bounded.
This can be problematic in imbalanced data distributions when samples from different categories are not equally distributed and a few clusters outnumber several rare clusters (which we will refer to as \emph{between cluster imbalanced}). In general, such highly skewed datasets impact the learning boundaries of cluster algorithms and can lead to a biased model, resulting in high-frequent, very large, or inequality scattered clusters. 
Another challenge which is not addressed properly by existing unsupervised clustering methods is \emph{within-cluster imbalance}, where a cluster is composed of several sub-clusters, and out-of-distribution samples. This can lead to the \emph{small disjuncts problem}~\cite{holte1989concept,rezaei2020generativecars} in which small disjuncts cover only a few samples. These samples usually have a much higher error rate compared to the samples from the same cluster.

\begin{figure} 
\includegraphics[width=0.5\textwidth]{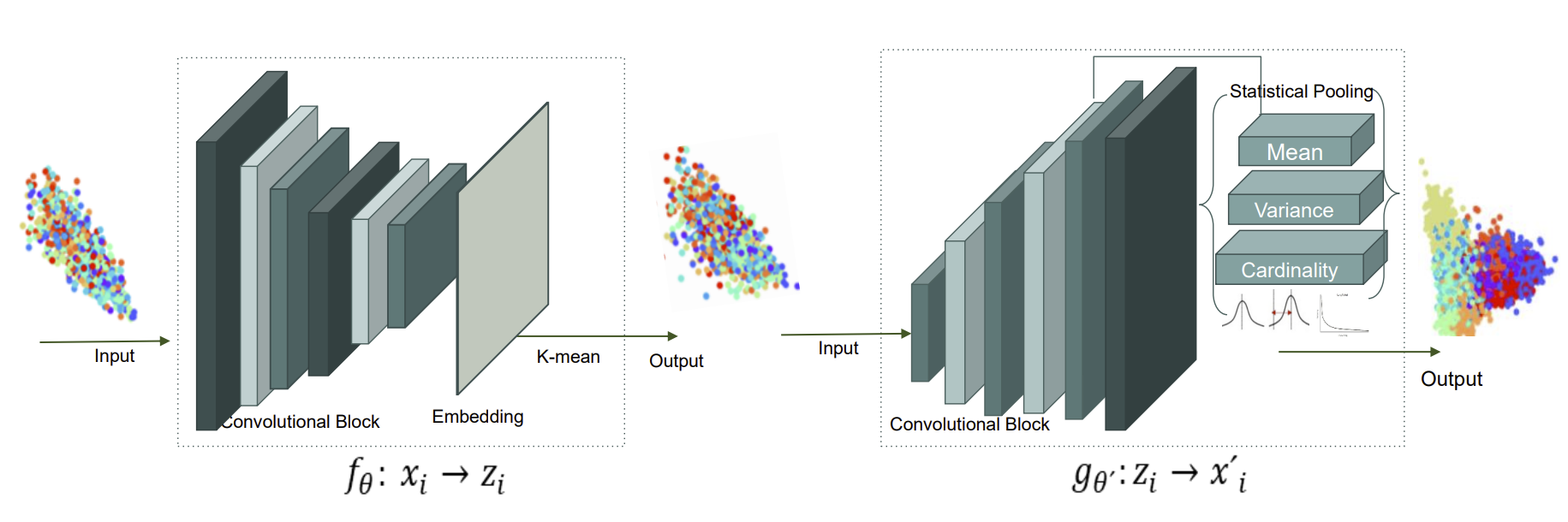}
\centering
\caption{Our proposed framework for simultaneous clustering and debiased representation learning. Our proposed method consists of an encoder clustering network (left part) and a decoder representation network (right part). The encoder takes a set of unlabeled images and performs clustering on top of the embedded layer while the decoder learns statistical representation, and performs reconstruction on the last layer.}
\label{fig_1}
\end{figure}

%Motivation of Statistical network
Most of the conventional clustering algorithms (e.g. K-means, self-organization map) do not provide a framework to tackle small disjuncts, out-of-distribution, or imbalanced distributions. However, the minority classes often contains very few instances with high degree of visual variability. Therefore, clusters with a large number of training instances may dominate the learning process during the inner-gradient steps, yielding low performance on clusters with fewer samples. To address these issues, we introduce a new unsupervised framework for joint debiased representation learning and image clustering. We develop a novel statistical pooling block that computes mean, variance, and estimates cardinality on the top of deep neural network's feature. The statistical pooling block alleviates the small-disjunct and imbalance problems. Furthermore, we refine the target distribution feature map by weighting less represented instances. This refined target distribution is also aware of cluster imbalances by taking cluster frequencies into account.

The main contributions in this work can be summarized as follows: 
\begin{itemize}
    \item We propose a novel end-to-end deep learning framework to jointly learn deep debiased representations and image clustering.
    \item We develop a statistical pooling block that learns to convey useful statistics about the data distribution such as mean, variance, and cardinality to effectively tackle challenges associated with out-of-distribution samples and data imbalance problem.
    \item We mitigate problems associated with small disjuncts and under-clustering by designing a new clustering loss function.
    \item We achieve comparable performance on imbalanced image clustering. Our experimental results show that the proposed framework achieves competitive performance on balanced image clustering and learns deep representations that can be transferred to other tasks and datasets.
\end{itemize}

%----------------------- RELATED WORK-----------------%
\section{Related Work} \label{related_work}
%This section describes related work carried out in the area of unsupervised clustering, joint representation learning and clustering, and learning representation of imbalanced data.

\textit{Clustering} Clustering has been broadly studied in machine learning in different aspects such as density-based clustering~\cite{guo2020density}, distribution-based clustering~\cite{huang2020deep,Alipour20}, grid-based clustering, distance-based clustering~\cite{Bicego20,cai2009locality,xing2003distance,dang2020multi}, grouping methods~\cite{zhuge2019simultaneous,WeiWYD020}. One of the most popular clustering methods is K-means~\cite{macqueen1967some}, which aims to partition the observation space into $k$ clusters so that each observation belongs to the cluster with the nearest centroid. Related ideas form the basis of a number of recent advancements, such as optimized K-means~\cite{wang2014optimized}, K-median~\cite{dasgupta2020explainable} or K-mean++~\cite{arthur2006k}. For a comprehensive literature survey of K-means related clustering algorithms, we refer readers to~\cite{patil2020survey}. In this paper, we a novel clustering loss function based on k-means on the top of features extracted by deep nezral networks.

\textit{Deep Clustering} Several approaches perform clustering on top of feature extracted by deep neural network~\cite{li2020deep,zhuge2019simultaneous,bengio2006greedy,chang2017deep}. Ji et al.~\cite{ji2019invariant} introduced IIC that trained deep neural networks for semantic clustering by maximizing mutual information between related pairs generated by random transforms. Deep embedded clustering~\cite{xie2016unsupervised} trains an auto-encoder with a reconstruction loss paired with a cluster assignment loss. It then defines a soft cluster assignment distribution by using k-means on top of the learned latent representations. The algorithm was later improved by an additional reconstruction loss to preserve local structure~\cite{guo2017improved}, an adversarial loss~\cite{mrabah2020adversarial}, an adversarial training procedure~\cite{tao2018rdec}, using data augmentation~\cite{guo2018deep}, using variational training~\cite{ghosh2019variational}, by adding a cluster frequency regularization~\cite{pmlr,ghasedi2017deep} or a confidence score~\cite{huang2020deep} to the clustering loss. We design a new clustering loss function to capture the under-clustering samples in imbalanced data distribution.

\textit{Joint Representation Learning and Image Clustering}
Recent studies~\cite{tao2018rdec,li2020deep,chen2020stochastic} have explored the combination of deep clustering with representation learning. Examples are,~\cite{yang2016joint} proposed JULE which jointly learns convnet features and clusters within a recurrent framework,~\cite{wu2019deep} proposed a deep comprehensive correlation mining (DCCM) that learns correlations behind the unlabeled data by supervision pseudo-label and triplet mutual information. \cite{caron2018deep} introduced an end-to-end training of visual features on a large dataset with a loss that attempts to preserve the information flowing through the reconstruction decoder network~\cite{guo2017improved,soleymani2022deep}. SCAN~\cite{van2020scan} proposed a two-step approach for image clustering. In first step, they use pretrained model and KNN while at the second step the clustering model is trained by imposing consistent predictions among neighbors. Another recent study, similar to our work, is deep fair clustering~\cite{li2020deep}, that aims to alleviate sensitive features during data partitioning by balancing the distribution of subgroups in each cluster. However, deep fair clustering only considers protected attributes such as skin color and does not consider statistical features and data distributions of each cluster. These and other previous studies such as SCAN~\cite{van2020scan}, IIC~\cite{ji2019invariant}, DCCM~\cite{wu2019deep} do not address problems arising due to small disjuncts, observations from imbalanced long-tailed distributions or out-of-distribution samples. 

\textit{Learning Imbalanced Long-tailed Distribution} 
%Since objects, events, and visual concepts appear with different frequencies in the real-world dataset, models have to be adapted to take these irregularities into account. 
Various approaches tackle learning imbalanced distribution in supervised deep learning either by modifying the data distribution or by defining a new cost-sensitive loss~\cite{cao2019learning,rezaei2020generative,rezaei2018generative}. Our work is similar to Task-Adaptive Meta-Learning (TAML)~\cite{lee2019learning, wang2021revisiting} which handles imbalance problems through statistical representations using meta-learning procedures. Here, we adapt this idea and transfer it to an unsupervised learning task. 

%----------------------- METHOD-----------------------%
\section{Method} \label{method}
Our goal is to learn unsupervised debiased representation and perform clustering. We first define notation and formulate a clustering task with our proposed method. Then we discuss the statistics pooling layer and deep representation network. %Next, we explain the optimization process and time complexity of our algorithm. 

We aim to cluster $N$ samples $\{x_i\}_{i=1}^N$ from the input space $\mathcal{X}=\mathbb{R}^{d_x}, d_x \in \mathbb{N}$ into $K$ clusters, represented by centroids $m_1,\ldots,m_K \in \mathbb{R}^{d_x}$. Our method includes two networks (see Fig.~\ref{fig_1}). The encoder network $f_{\theta}: x_i\rightarrow z_i$ maps an input image $x_i \in \mathcal{X}$ to its latent embedding $z_i\in\mathcal{Z}$. The decoder network $g_{\theta'}:z_i\rightarrow x_i^\prime$ reconstructs $x_i$ from its latent embedding $z_i$. Similar~\cite{xie2016unsupervised}, the networks are first trained jointly as a standard autoencoder, with reconstruction loss $\mathcal{L}_r$ to minimize the mean squared difference between each $x_i$ and $x_i^\prime$. After convergence, the autoencoder is then fine-tuned with a combined loss function (Eq.\ref{eq_1}) consisting of the autoencoder reconstruction loss and a clustering loss $\mathcal{L}_c$. The relative weight of each of the two losses is indicated by $\lambda \in (0,1)$, which controls the degree of distortion introduced in the embedded space:

\begin{equation} \label{eq_1}
\mathcal{L} =  \lambda \mathcal{L}_{c} + (1-\lambda) \mathcal{L}_{r}.
\end{equation}

\noindent We follow the suggestion of \cite{guo2017improved} for the configuration of $\mathcal{L}$ and set ${\lambda}$ to 0.1 (if not stated otherwise). 

\subsection{Deep Clustering and Parameter Initialization} \label{method-init}
After convergence of the first training step of our network, yielding a good embedding representation of each training sample in the first step, we perform clustering in the latent space $\mathcal{Z}$ using Kullback-Leibler (KL) divergence loss:

\begin{equation} \label{eq_3}
\mathcal{L}_c=\text{KL}(P||Q) =
\sum _{i} \sum _{j} p_{ij}\ln {\frac {p_{ij}}{q_{ij}}}
\end{equation}

\noindent where $Q$ is a soft labeling distribution with elements $q_{ij}$. $P$ is an auxiliary target distribution derived from $Q$ with elements $p_{ij}$. More specifically, $p_{ij}$ are the elements of the target distribution while $q_{ij}$ is the distance between the embedded $z_i$ and the center $m_j$ of the $j$-th cluster. This distance is measured by a Student's \textit{t-}distribution (cf. ~\cite{van2009learning}):

\begin{equation} \label{eq_4}
\begin{split}
q_{ij} = \frac{
    (1+ \norm{z_i - m_j}^2 / \alpha )^{-\frac{\alpha +1}{2}}
}{
    \sum_{k} \left(1+ \norm{z_i - m_k}^2 / \alpha \right)^{-\frac{\alpha +1}{2}}
},
\end{split}
\end{equation}

\noindent where $\alpha$ is the degrees of freedom of the Student’s \textit{t-}distribution (we here only consider $\alpha=1$).

To address problems associated with small disjuncts, we modify the target distribution $P$ by pushing data points that are similar in the original space closer together in the latent space. Thereby samples from less-frequent classes can be identified as a cluster. Then, $p_{ij}$ is computed as follows:

\begin{equation} \label{eq_5}
p_{ij} = \frac{q_{ij}^2 / (u_{j} + v_{j} )}{\sum_k q_{ik}^2 / v_{k}}
\end{equation}

\noindent where $u_{j} = \sum_i q_{ij}$ are soft cluster frequencies while $v_{j}$ normalize the frequency of samples per cluster. We enforce inter-cluster margins irrespective of different cluster sizes and variations. This difference leads to the unique capability in preserving discrimination in small disjunct and forming a local clustering boundary that is insensitive to imbalanced cluster sizes. The sample's frequency calculated as: 

\begin{equation} \label{eq_5_2}
v_{j} = - \sum_i \sum_j \sqrt{ \frac{ \sum_k N_k}{N_j} {(1-q_{ij})}^\gamma \log (q_{ij})}.
\end{equation}

\noindent Here, $N_j$ is the estimated cardinality of cluster $j$, $\gamma$ is a relaxation parameter in laymen’s terms and set to 2. Note that, to prevent instability in the training procedure, we do not update $P$ at every iteration. $P$ is only updated if changes in the label assignments between two consecutive updates of the target distribution are less than a threshold $\delta$. This tolerance threshold and its empirical property are discussed in more detail in Section~\ref{Experiments}.

For parameter initialization, we follow the standard procedure by \cite{xie2016unsupervised} and \cite{guo2017improved}: the autoencoder is pre-trained separately, and the centroids $m_1,\ldots,m_K$ are initialized by performing standard $K$-means clustering on the latent embeddings of the training samples.

\subsection{Deep Debiased Representation Learning and Reconstruction Loss}

The embedded feature space of the unsupervised encoder alone does not yield good clustering performance when the data distribution is imbalance. Deep embedded clustering optimizes cluster assignments by forcing data around the centroid in the bottleneck where samples from the less frequent cluster have less chance to be assigned as a centroid. We use statistics pooling layer to address this issue and improve the clustering performance by manipulating and balancing the decoder feature space based on a normalization of the previously learned feature vector (see Eq.(\ref{eq_5}) and Eq.(\ref{eq_5_2})). 

\subsubsection{Statistics Pooling Layer}

The statistics pooling refines the latent features from the encoder parameters ($\theta$) into a more informative representation for clustering imbalanced datasets and out-of-distribution samples. From the encoder clustering loss and soft label assignment, we obtain the approximate number of clusters as well as the number of samples at each cluster. Consider the cardinality $N_k$ be the set of samples that belong to the cluster $k \in \{k = 1, . . . , K\}$. We expect the cardinality $N_k$ to be large for long-tailed distribution with a small number of training samples. Each sample in the set with a similar label is transformed by shared non-linearity and then averaged to create a vector ($\mu$). To prevent the procedure to output statistics based on single instances, we compute the variance ($\sigma$) as a second vector.  %= \frac{1}{c} {\sum\limits_{i=1}^N {(x_{i} - \mu_\beta})^2}$
The result of our statistics pooling is given by concatenation of all statistics (cardinality, mean, and variance): 

\begin{equation} \label{eq_6}
S_{\theta} (x) =  (N_k, \mu (x), \sigma (x)).
\end{equation}

\noindent Note that $\mu$ and $\sigma$ are calculated on the samples that belong to the same cluster. %More details could be found in our open-source implementation~\footnote{https://we.tl/t-ADmGWuGGYk}. %The Algorithm~\ref{alg2} summarizes the training procedure.

\subsection{Optimization}

We perform multi-objective optimization to jointly optimize the cluster loss and the reconstruction loss using mini-batch stochastic gradient descent (SGD). In each iteration the clustering network’s weights~$\theta$, cluster centers~$m_j$, statistics decoder's weights~${\theta}'$, and target distribution~$P$ are updated and optimized on the basis of \eqref{eq_1}. %The problem of Eq. \ref{eq_1}, is not convex over all of variables.
The gradient of $ \lambda \mathcal{L}_{c}$ with respect to the embedded points $z_i$ and cluster centers $m_j$ for fixed target distribution $P$ are calculated as:

\begin{equation} \label{eq_7}
\frac{\partial \mathcal{L}_{c}}{\partial z_i} = 2 {\sum\limits_{j=1}^K {(1+ \norm{z_i - {m}_j}^2 )^{-1}  (p_{ij} - q_{ij}) (z_i - m_j)}};
\end{equation}

\begin{equation} \label{eq_8}
\frac{\partial \mathcal{L}_{c}}{\partial m_j} = 2 {\sum\limits_{i=1}^n {(1+ \norm{z_i - {m}_j}^2 )^{-1}  (q_{ij} - p_{ij}) (z_i - m_j)}}.
\end{equation}

\noindent The gradients ${\partial \mathcal{L}_{c}} / {\partial z_i}$ are directly used in the backpropagation step. We further use \eqref{eq_8} to update the cluster centers $\mu_j$ as follow: 

\begin{equation} \label{eq_9}
{m}_j = {m}_j - \frac{\eta}{m} \sum\limits_{i=1}^n  \frac{\partial \mathcal{L}_{c}}{\partial m_j}, 
\end{equation}

\noindent where $\eta$ is a learning rate and $n$ is the size of the mini-batch. The network's weights of the statistics decoder are updated as Eq.~\ref{eq_10}.

\begin{equation} \label{eq_10}
{\theta}' = {\theta}' - \frac{\lambda}{n} \sum\limits_{i=1}^n  \frac{\partial \mathcal{L}_{r}}{\partial {\theta}'}.
\end{equation}

\noindent Note that in~\eqref{eq_10} $\mathcal{L}_{r} =  \sum\limits_{i=1}^n \norm{x_i - g_{{\theta}'} (z_i)}^2$
and ${\partial \mathcal{L}_{r}} / {\partial {\theta}'}$ is calculated by:

\begin{equation} \label{eq_12}
\frac{\partial \mathcal{L}_{r}}{\partial {\theta}'} = 2 {\sum\limits_{j=1}^K {(1+ \norm{z_i - {m}_j}^2 )^{-1}  (p_{ij} - q_{ij}) (z_i - m_j)}}.
\end{equation}

\noindent Finally, the statistics encoder weights are updated by:

\begin{equation} \label{eq_13}
{\theta} = {\theta} - \frac{\lambda}{n} \sum\limits_{i=1}^n  \left( \frac{\partial \mathcal{L}_{r}}{\partial {\theta}} +  \frac{\partial \mathcal{L}_{c}}{\partial {\theta}} \right) 
\end{equation}

\noindent with

\begin{equation} \label{eq_14}
\begin{split}
 \frac{\partial \mathcal{L}_{r}}{\partial {\theta}} &=  \frac{\partial}{\partial {\theta}} \frac{1}{2n} \sum\limits_{i=1}^n ({x_i - f_{\theta} (z_i)})^2 \\
 &= \frac{1}{2n} \sum\limits_{i=1}^n \frac{\partial}{\partial {\theta}} ({x_i - f_{\theta} (z_i)})^2 \\
 &= \frac{1}{n} \sum\limits_{i=1}^n  (x_i - f_{\theta}) \frac{\partial}{\partial {\theta}} (x_i - f_{\theta}) \\
% = \frac{1}{2n} \sum\limits_{i=1}^n (x_i - f_{\theta}) \left( \frac{\partial}{\partial {\theta}} x_i - \frac{\partial}{\partial {\theta}} f_{\theta} \right)\\
 &= \frac{1}{n} \sum\limits_{i=1}^n  (x_i - f_{\theta}) \frac{\partial}{\partial {\theta}} f_{\theta}.
\end{split}
\end{equation}

\noindent By iterating these updates, the label assigned of $x_i$ is obtained using $y_i =   \underset{j}{\mbox{argmax}} \,  q_{ij}$ where $q_{ij}$ is computed as described in Eq.~\ref{eq_4}. The training process is repeated until a convergence criterion based on the KL loss is met. 

%\begin{equation} \label{eq_15}
%y_i =   \underset{j}{\mbox{argmax}} \,  q_{ij},
%\end{equation}

%where $q_{ij}$ is computed as described in Eq.~\ref{eq_4}. The training process is repeated until a convergence criterion based on the KL loss is met. 

The time complexity of our algorithm where the network architecture included $D$ neurons in hidden layers can be $O(nD^2)$.

%----------------------- EXPERIMENTS------------------%
\section{Experiments} \label{Experiments}
In this section, we conduct several experiments and ablation analysis to examine our proposed framework. First, we compare our achieved performance with some recent related works for the task of clustering of uniform distribution as well as imbalanced distribution. Second, we study the impact of our proposed method for learning out-of-distribution samples and the ability to handle the imbalanced data situation. 

%-----------------------------------------------------------------------------
%-----------------------------------------------------------------------------

%---------------------------------------------------
%     Table Imbalanced CIFAR 10, CIFAR-100
%---------------------------------------------------
\begin{table*} [h]
\centering
\caption{Comparison results of our achieved accuracy on imbalanced CIFAR-10 and CIFAR-100.}
\label{tableimbCIFAR}
\begin{tabular}{l l l l l l l l l }
\toprule
Dataset  & \multicolumn{4}{c}{Imbalanced CIFAR-10} &  \multicolumn{4}{c}{Imbalanced CIFAR-100} \\
\midrule
Imbalanced Type & \multicolumn{2}{c}{long-tailed} & \multicolumn{2}{c}{step}  & \multicolumn{2}{c}{long-tailed}  & \multicolumn{2}{c}{step} \\
\midrule
Imbalanced Ratio & 10 & 100 & 10 & 100 & 10 & 100 & 10 & 100  \\
\midrule
\multicolumn{9}{c}{Unsupervised Learning Method}  \\
\midrule
IDEC~\cite{guo2017improved} & 0.2533 & 0.1703 & 0.2604 & 0.2136 & 0.1024 & 0.1053 &  0.1280& 0.1301\\
DEC~\cite{xie2016unsupervised} & 0.2061 & 0.1288 & 0.2136 & 0.1645 & 0.0848 & 0.0740 & 0.1402 & 0.1276\\
VAE & 0.2173 & 0.1362 & 0.2071 & 0.1753 & 0.0963 & 0.0821 & 0.1637 & 0.1351\\
DAC~\cite{chang2017deep} & 0.2965 & 0.2874 & 0.2717 & 0.2241 & 0.1886 & 0.1255 & 0.2316 & 0.2471\\
DCCM~\cite{wu2019deep} & 0.2807 & 0.2663 & 0.243 & 0.2091 & 0.1814 & -  & - & - \\
SCAN~\cite{van2020scan} & 0.4052 & 0.3431 & 0.3478 & {0.5057} & 0.2016 & - & - & -\\
IIC~\cite{park2020improving} & 0.3269 & 0.3028 & 0.2856 & 0.3577 & \textbf{0.2512} & 0.1172 & 0.2771 & 0.2943 \\
our method (statDEC)  & \textbf{0.4831} & \textbf{0.4106} & \textbf{0.4872} & \textbf{0.4511} & 0.2365 & \textbf{0.1843} & \textbf{0.3716} & \textbf{0.3502} \\
\midrule
\multicolumn{9}{c}{Supervised Learning Method}  \\
\midrule
TAML~\cite{lee2019learning} & \underline{0.7733} & \underline{0.7209} & \underline{0.7485} & \underline{0.7124} & \underline{0.7225} & \underline{0.7128} & \underline{0.7320} & \underline{0.7286}\\
LDAM~\cite{cao2019learning} & 0.6313 & 0.6235 & 0.7014 & 0.6521 &  0.4211 & 0.4241 & 0.4656 & 0.6056\\
Focal Loss~\cite{lin2017focal} & 0.4089 & 0.2959 & 0.6062 & 0.6521 & 0.4654 & 0.3806 & 0.4422 & 0.4058\\
\bottomrule
\end{tabular}
\end{table*}

\paragraph{Dataset} The proposed method is evaluated on MNIST~\cite{lecun1998gradient}, CIFAR-10, CIFAR-100~\cite{krizhevsky2009learning}, imbalanced CIFAR-10, imbalanced CIFAR-100 with two different imbalanced ratio, SVHN~\cite{netzer2011reading}, and a real-world medical imaging dataset REFUGE-2~\cite{orlando2020refuge}.  

\textit{MNIST} consists of 60,000 images for training and 10,000 for testing, each image has a size of $28 \times 28$ pixels and is from one of 10 classes. We train on the full training set and report the results on the test set.

\textit{Imbalanced CIFAR-10 and CIFAR-100} 
CIFAR-10 and CIFAR-100 are subsets of the tiny images dataset. Both datasets include 50,000 images for training and 10,000 validation images of size $32 \times 32$ with 10 and 100 classes, respectively. We create the imbalanced version of these datasets by reducing the number of examples per class with two different imbalance ratios and two different types of imbalances: the step imbalance~\cite{buda2018systematic} and long-tailed imbalance~\cite{cui2019class} configuration. For the CIFAR dataset with step imbalance, all minority classes have the same number of examples, so have the majority classes. The imbalance ratio computes the number of samples of the minority class divided by the number of samples of the majority class. The long-tailed imbalance distribution follows an exponential decay of classes, consisting of ahead of the distribution (majority class) and the long-tailed minority classes with different numbers of samples.

\textit{SVHN}: consists of 26,032 images with size of $32 \times 32$ from 10 digits classes.

\textit{REFUGE-2}~\cite{orlando2020refuge} is a public challenge and part of the MICCAI-2020 conference. The organizers released 1,200 microscopy retinal scans with a size of $2124 \times 2056$ pixels from two different machines and scanned by two clinics. The dataset is imbalanced with a ratio of 1:30.  

\paragraph{Evaluation Metrics} As an unsupervised evaluation metric, we use the clustering Accuracy (ACC), Normalized Mutual Information (NMI), and Adjusted Rand Index (ARI) for evaluations. These measures have values in $[0, 1]$, higher scores show more accurate clustering results.

\paragraph{Compared Methods} We compare our results with unsupervised DEC~\cite{xie2016unsupervised}, IDEC~\cite{guo2017improved}, and VDEC~\cite{ghosh2019variational}. These methods can be viewed as a variant of our method when the reconstruction loss and network architecture are different. Note that the reported results for DEC and IDEC are based on our implementation and the results for SCAN~\cite{van2020scan}, IIC~\cite{ji2019invariant}, and DCCM~\cite{wu2019deep} are based on GitHub code by the authors. We further demonstrate the effectiveness of the proposed method in handling out-of-distribution and imbalanced data situations by comparing results with and without statistical pooling layer and contrast results to those of two supervised state-of-the-art techniques LDAM~\cite{cao2019learning} and TAML~\cite{lee2019learning}.

%-----------------------------------------------------------------------------
%-----------------------------------------------------------------------------
\paragraph{Experimental Setting}\label{discussion} Our implemented method, \emph{StatDEC}, includes a fully-connected multi-layer perceptron (MLP) with dimensions \textit{$d_x$-500-500-1000-10} as encoder for all balanced and imbalanced datasets. Here, $d_x$ is the dimension of the input data. The decoder network is also a fully-connected MLP with dimensions \textit{10-1000-500-500-$d_x$}. Our networks contain one statistical pooling layer, on top of the second hidden layer of the decoder network. The target distribution is weighted according to Eq.~\ref{eq_5}. Each layer is pre-trained for 100,000 iterations with dropout. The entire deep autoencoder is further fine-tuned for 200,000 iterations without dropout for both layer-wise pre-training and end-to-end tuning. The minibatch size is set to 256 for MNIST, 128 for CIFAR, and 8 for REFUGE. We use a learning rate of 0.01 which is divided by 10 every 20,000 iterations and set weight decay to zero. After pretraining, the coefficient $\lambda$ of clustering loss is set to 0.1. The convergence threshold $\delta$ is set to 0.001 while the update intervals for target distribution are 70, 80, 100, 120 iterations for REFUGE, MNIST, CIFAR-10, and CIFAR-100 respectively.

%-----------------------------------------------------------------------------
\paragraph{Unsupervised clustering} 
We validate the effectiveness of our method on both uniform and long-tailed datasets, following the experimental method in~\cite{cao2019learning} on creating imbalanced CIFAR, we sampled subsets of CIFAR-10 and CIFAR-100 with two different imbalanced ratios of 10 and 100. 
Table~\ref{tableimbCIFAR} summarizes the quantitative results for the step imbalanced and long-tailed imbalanced CIFAR-10 and CIFAR-100 data. Our configuration with statistical pooling and weighted target distribution (StatDEC) achieves 35\% accuracy, the highest accuracy for the step imbalanced CIFAR-100 data. One reason for this improved performance over other approaches is that we consider cluster sizes in the computation of the target distribution. In contrast, DEC and IDEC are independent of cluster sizes and all label assignments are based on distances between the embedded space and centroids.

Compared to state-of-the-art supervised methods we find that TAML~\cite{lee2019learning} achieves the best performance among the supervised model. TAML addressed the imbalanced problem using two statistics pooling layers with a meta-learning procedure. The comparison of SCAN~\cite{van2020scan} and StatDEC in Table~\ref{tableimbCIFAR} shows that SCAN achieved best performance on step imbalanced CIFAR-10. The SCAN loss enforces consistent prediction among neighbors and it maximizes entropy to avoid all samples being assigned to the same cluster. However, It's important to mention the SCAN paper is trained with ResNet-50 architecture while our encoder architecture is a simple multi-layers preceptron.  

The points that can be inferred from Table~\ref{tableimbCIFAR} are: (1) The achieved accuracy by representation-based clustering algorithms (i.e. AE~\cite{bengio2006greedy}) is higher than traditional clustering techniques (i.e. K-means~\cite{wang2014optimized}). This shows the importance of representation learning on image clustering. (2) Best performing algorithms for image clustering simultaneously perform representation learning and clustering (i.e. SCAN~\cite{van2020scan} and JULE~\cite{yang2016joint}).

\paragraph{Handling small disjunct problem}~\label{section4.5.2}
We validate the effectiveness of our model for handling the small disjunct problem on real-world clinical datasets released by the REFUGE-2 challenge~\cite{orlando2020refuge}. The REFUGE 2020 training dataset comprises 400 subjects with glaucoma or non-glaucoma disease. The non-glaucoma category includes healthy or diseases other than glaucoma. Due to this fact, the clustering algorithm may detect more than two clusters. To prevent this problem, we proposed to weight the target distribution to push data points that are similar in the original space closer together in the latent space. We attended the REFUGE challenge with our proposed StatDEC model and achieved an area under the ROC curve (AUC) of \textbf{0.7865} as reported by the challenge's organizer~\footnote{https://refuge.grand-challenge.org/validationleaderboard/} on the 400 testing subjects. 

\begin{figure}[!t]
  \centering
  \subfloat[]{\includegraphics[width=0.40\textwidth]{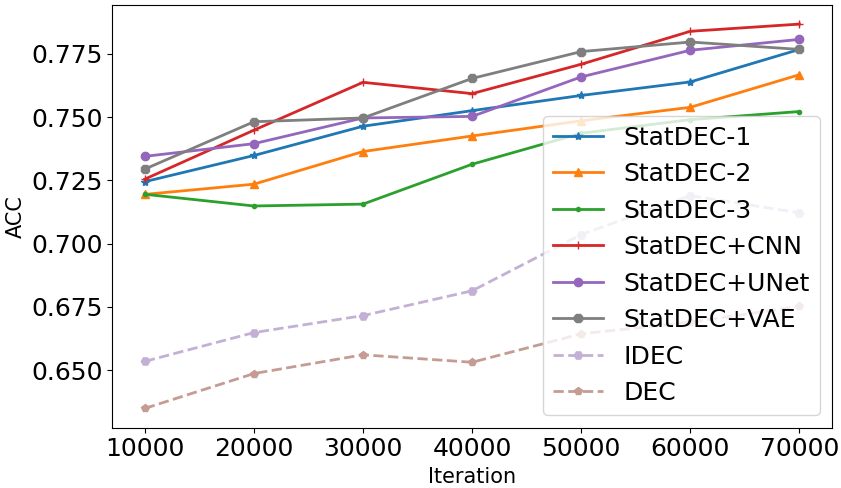}\label{figr_1a}}
  \hfill
  \subfloat[]{\includegraphics[width=0.40\textwidth]{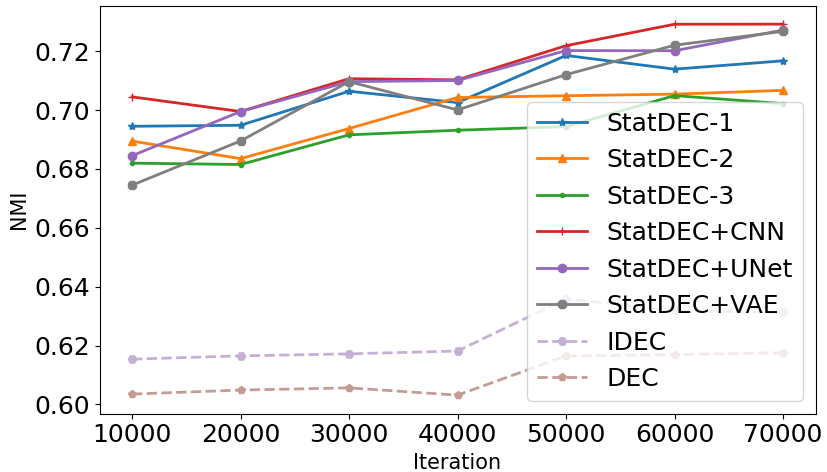}\label{figr_2b}}
   \hfill
  \subfloat[]{\includegraphics[width=0.40\textwidth]{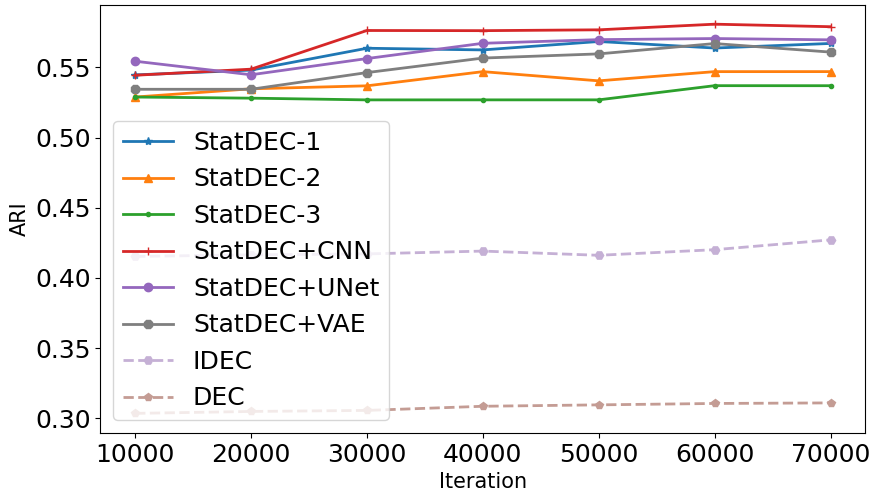}\label{figr_3c}}
   \hfill
  \subfloat[]{\includegraphics[width=0.38\textwidth]{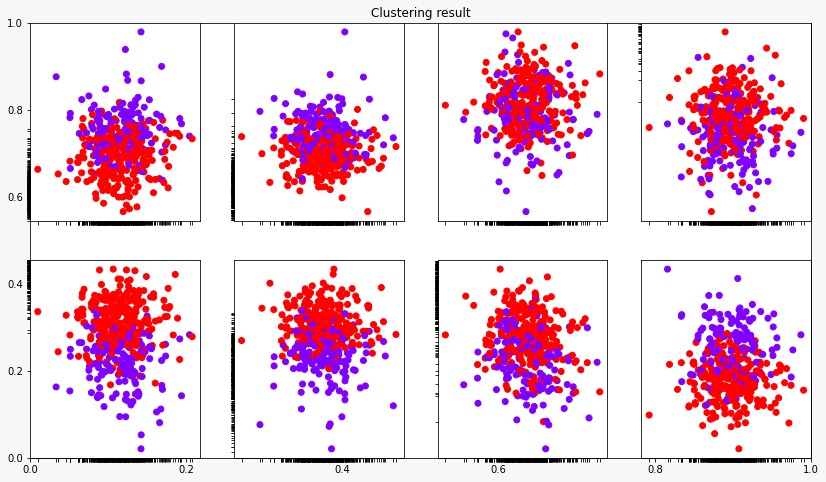}\label{figr_4d}}
  \caption{(a) Comparison of clustering performance on REFUGE dataset in terms of accuracy (a), NMI (b), and ARI (c). Visualization of deep embedded on REFUGE dataset (d).}
\label{fig_refuge}
\end{figure}

Having no access to the test and validation labels, we split the training data into 300 subjects for training and 100 as local testing set to report performance measures and clustering results. Figure~\ref{fig_refuge} compares clustering results on this local test set in terms of accuracy (\ref{figr_1a}), NMI~(\ref{figr_2b}) and ARI~(\ref{figr_3c}).  

\section{Ablation Analysis}

We train different configuration of our proposed model to answer the following questions: \textbf{Q1:} How do statistical pooling impact \textit{imbalanced treatment}, \textit{out-of-distribution problem}, and \textit{small disjunct} problem? 
\textbf{Q2:} What is the effect of balancing the target distribution on imbalanced treatment? \textbf{Q3:} How our contributions affected by different datasets and different network architecture? \textbf{Q3:} How our proposed method works on balanced dataset? 

\begin{figure}[!h]
  \centering
      \subfloat[]{\includegraphics[width=0.437\textwidth]{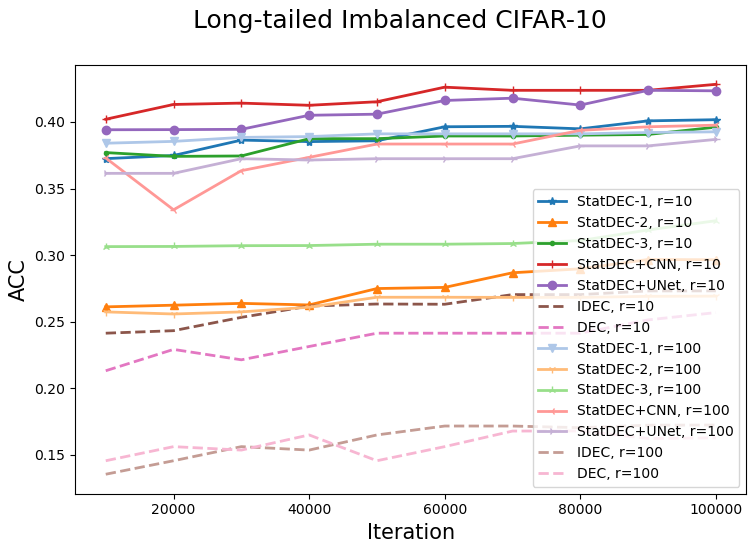}\label{figimb_a}}
        \hfill
  \subfloat[]{\includegraphics[width=0.44\textwidth]{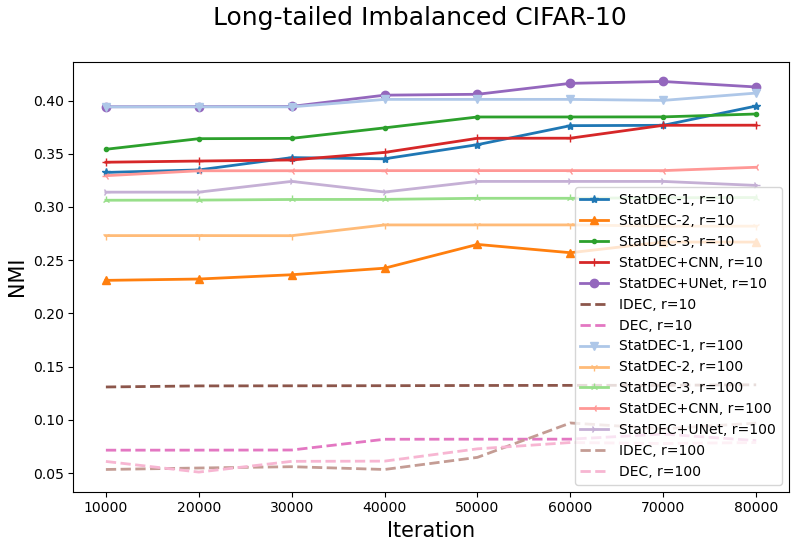}\label{figimb_b}}
    \caption{Comparison of clustering performance on long-tailed imbalanced CIFAR-10 in terms of accuracy (a) and NMI (b). Visualization results of embedded clustering obtained by StatDEC-1 on step-imbalanced CIFAR-10 with a ratio of 10. (c). Visualization of deep embedded clustering on long-tailed imbalanced CIFAR-10 with a ratio of 100. (d)}
\label{fig_imb}
\end{figure}

Based on evaluation results in terms of accuracy, adjusted rand index (ARI), and normalized mutual information (NMI) in Table~\ref{tableimbCIFAR} and Figure~\ref{fig_imb} confirm the success of our methods for imbalanced treatment. We performed several experiments considering different imbalanced ratios and different imbalanced types to compare and validate our results with other clustering algorithms and supervised methods (see Table~\ref{tableimbCIFAR}).

To validate our developed statics pooling for handling small disjunct problems, we trained our models with a REFUGE dataset. The retina images in this dataset are collected from people of different nationalities in two hospitals and the probability of having a disjunct is high where a small disjunct covers only a few training examples. The qualitative results are depicted in Figure~\ref{fig_refuge}.

We perform additional experiments to study the ability to generalize beyond the biases of a training set. We trained our models on cropped SVHN~\cite{netzer2011reading} dataset and tested them on MNIST~\cite{lecun1998gradient} to validate it for handling out-of-distribution at testing time. Table~\ref{tableood} shows and compares the achieved accuracy on resized MNIST dataset when the model trained only with SVHN samples. It can be inferred from Table~\ref{tableood}, the trained networks included statics pooling (StatDEC, StatDEC-3) are more successful for training bias and handling distributional shift. The objective here is to show the ability to cluster out-of-distribution samples.

%---------------------------------------------------
%     Table MNIST,  CIFAR 10, CIFAR-100
%---------------------------------------------------
\begin{table*} [!h]
\centering
\caption{Comparison results of our achieved accuracy on MNIST, CIFAR-10, and CIFAR-100. Results of (*) methods are taken from ~\cite{chang2017deep} and results of (+) is based on CIFAR-20.}
\label{tableCIFAR}
\begin{tabular}{l l l l l l l l l l }
\toprule
Model  & \multicolumn{3}{c}{MNIST~\cite{lecun1998gradient}} &  \multicolumn{3}{c}{CIFAR-10~\cite{krizhevsky2009learning}} &  \multicolumn{3}{c}{CIFAR-100~\cite{krizhevsky2009learning}} \\
\midrule
  & NMI & ARI & ACC & NMI & ARI & ACC & NMI & ARI & ACC \\
\midrule
%StatDEC-2 & 0.8074 & 0.7927 & 0.8526 & 0.2965 & 0.2514 & 0.4134 & 0.1863 & 0.1048 & 0.2457 \\
%StatDEC-3 & 0.8622 & 0.8445 & 0.9214 & 0.3961 & 0.3358 & 0.5078 & 0.2876 & 0.2380 & 0.3409 \\
%StatDEC+CNN & 0.8905 & 0.8756 & 0.9368 & 0.4227 & 0.3976 & 0.5184 & 0.3175 & 0.2176 & 0.3638 \\
%StatDEC+UNet & 0.9066 & 0.8826 & 0.9407 & 0.4234 & 0.3869 & 0.5227 & 0.3084 & 0.2293 & 0.3571 \\
%StatDEC+VAE & 0.9041 & 0.8743 & 0.9321 & 0.4106 & 0.3451 & 0.5130 & 0.2945 & 0.2037 & 0.3440 \\
*K-means \cite{wang2014optimized} & 0.4997 & 0.3652 & 0.5723 & 0.0871 & 0.0487 & 0.2289 & 0.0839 & 0.0280 & 0.1297 \\
*NMF~\cite{cai2009locality} & 0.6082 & 0.4298 & 0.5447 & 0.0814 & 0.0338 & 0.1895 & 0.0791 & 0.0263 & 0.1175\\
AE~\cite{bengio2006greedy} &  0.7257 & 0.6139 & 0.8123 & 0.2393 & 0.1689 & 0.3135 & 0.1004 & 0.0476 & 0.1645 \\
VAE~\cite{kingma2013auto} & 0.7364 & 0.7129 & 0.8317 & 0.2451 & 0.1674 & 0.2908 & 0.1079 & 0.0403 & 0.1517 \\
*JULE~\cite{yang2016joint} & 0.9130 & 0.9270 & 0.9640 & 0.1923 & 0.1377 & 0.2715 & 0.1026 & 0.0327 & 0.1367 \\
IDEC ~\cite{guo2017improved}& 0.8169 & 0.8687 & 0.8806 & 0.2731 & 0.1723 & 0.3169 & 0.1407 & 0.0429 & 0.1911 \\
DEC~\cite{xie2016unsupervised}& 0.7716 & 0.7414 &0.8430& 0.2568 & 0.1607 & 0.3010 & 0.1358 & 0.0495 & 0.1852\\
VDEC~\cite{ghosh2019variational} & 0.8364& 0.7482 & 0.8426 & 0.4151 & 0.2674 & 0.3908 & 0.1925 & 0.1458 & 0.2190  \\
*DAC~\cite{chang2017deep} & \textbf{0.9351} & \textbf{0.9486} & \textbf{0.9775} & 0.3793 & 0.2802 & 0.4982 & 0.1623 & 0.1776 & 0.3189 \\
DCCM~\cite{wu2019deep} & - & - & - & 0.4961 & 0.4082 & 0.6253 & 0.2625 & 0.1733 & 0.3271 \\
IIC~\cite{ji2019invariant} & - & - & 0.9923 & 0.5141 & 0.4118 & 0.6174 & 0.2251 & 0.1170 & 0.2574 \\
+SCAN~\cite{van2020scan} & - & - & - & \textbf{0.7976} & \textbf{0.7724} & \textbf{0.8836} & - & - & - \\
StatDEC & 0.8931 & 0.8764 & 0.9343 & 0.4017 & 0.3926 & 0.5283 & \textbf{0.2697} & \textbf{0.2013} & \textbf{0.3501} \\
\bottomrule
\end{tabular}
\end{table*}

%---------------------------------------------------
%     Table OOD
%---------------------------------------------------

\begin{table} [!htbp]
\centering
\caption{Ablation study on distributional shift at test time: Comparison results of our methods on MNIST dataset when the network trained on SVHN samples.}
\label{tableood}
\begin{tabular}{l l l l }
\toprule
Trained Dataset  & \multicolumn{3}{c}{SVHN}  \\
\midrule
Measurements & NMI & ARI & ACC  \\
\midrule
StatDEC  & 0.7305 & 0.6921 & 0.8236  \\
StatDEC-2  & 0.5829 & 0.5415 & 0.6810 \\
StatDEC-3  & 0.7083 & 0.6711 & 0.8174 \\
IDEC~\cite{guo2017improved} & 0.3721 & 0.2948 & 0.5104 \\
DEC~\cite{xie2016unsupervised} & 0.2906 & 0.3208 & 0.4836 \\
\bottomrule
\end{tabular}
\end{table}

To answer Q3, we implemented StatDEC with the different base networks. We evaluate three architectural choices for our proposed method by using different types of autoencoders. We consider an autoencoder based on a deep neural network (StatDEC, StatDEC-2, StatDEC-3), a convolutional autoencoder (StatDEC+CNN , StatDEC+UNet), and a variational autoencoder (StatDEC+VAE). Table~\ref{tablearchitrcture} shows different configurations.

%---------------------------------------------------
%     Table Architecture
%---------------------------------------------------
\begin{table} [!t]
\centering
\caption{Network Architecture.}
\label{tablearchitrcture}
\begin{tabular}{l l l l }
\midrule
Our Model &  Weighted Loss & StatPooling & Network  \\
\midrule
StatDEC-1  & yes & yes & DNN  \\
StatDEC-2  & yes & no  & DNN  \\
StatDEC-3  & no  & yes & DNN  \\
StatDEC+CNN  & no  & yes & CNN  \\
StatDEC+UNet  & yes & yes & UNet \\
StatDEC+VAE  & yes & yes & VAE \\
\bottomrule
\end{tabular}
\end{table}

StatDEC-2 shows the impact of statistical pooling where it has similar architecture and configuration as StatDEC, but without statistical pooling. The target distribution is weighted as described in Section~\ref{method}. The \textit{StatDEC-3}, presents the impact of modifying the target distribution with cluster frequency and sample frequency. The decoder network includes a statistical pooling layer. 

In a different experiment, we study the impact of the network architecture where \textit{StatDEC+CNN} is a convolution encoder architecture and composed of four convolutional layers with batch normalization and ReLU activation for each layer, followed by a flattening operation and a fully connected layer on top of the convolutional layers. The decoder convolution is composed of a fully connected layer and three transposed convolutions with batch normalization and ReLU activation in between. All convolutions and transposed convolutions use a filter size of $4 \times 4$. All other parameters are defined as in StatDEC.

The \textit{StatDEC+UNet} network has a similar architecture to UNet~\cite{ronneberger2015u}, including skip connection layers in the decoder network. Each skip connection concatenates all channels at layer $i$ with those of layer $n-i$, where $n$ is the total number of layers. We use convolutional layers with kernel size 5 $\times$ 5 and stride 2 in the encoder part. In the decoder, we perform up-sampling by image re-size layers with a factor of 2 and a convolutional layer with kernel size 3 $\times$ 3 and stride 1.

In \textit{StatDEC+VAE} network, we experiment the impact of the variational encoder as a network for image clustering and probabilistic decoder as a network for robust statistical representation learning. Here, instead of encoding an input as a single point, we encode it as a distribution over the latent space. First, the input data $x$ is encoded as distribution over the latent space $p(z|x)$ and second, $z$ sampled the latent representation $z \sim p(z|x)$. Then, the sampled point is decoded and the reconstruction error can be computed. The probabilistic decoder network includes a statistical pooling layer on top of the second hidden layer. We perform image clustering on top of deep variational embedded. During optimization, variation encoder's weights, decoder's weights, cluster centers, and target distribution are updated. 

Figure~\ref{figimb_a}, compares the impact of different architecture on imbalanced CIFAR-10 and CIFAR-100.

% ----------------
% CIFAR10
%-----------------

%\begin{figure} [!h]
%  \centering
%   \subfloat[]{\includegraphics[width=0.46\textwidth]{ICCV/figures/CIFAR10_ACC.png}\label{figcifar10_a}}
%        \hfill
%   \subfloat[]{\includegraphics[width=0.46\textwidth]{ICCV/figures/CIFAR10_NMI.png}\label{figcifar10_b}}
%           \hfill
%   \subfloat[]{\includegraphics[width=0.46\textwidth]{ICCV/figures/CIFAR10_ARI.png}\label{figcifar10_c}}
%  \caption{Comparison of clustering performance on CIFAR-10 in terms of accuracy (a), NMI (b), and ARI (c).}
%\label{fig_cifar10}
%\end{figure}

% ----------------
% CIFAR100
%-----------------

%\begin{figure} [!h]
%  \centering
%   \subfloat[]{\includegraphics[width=0.46\textwidth]{ICCV/figures/CIFAR100_ACC.png}\label{figcifar100_a}}
 %       \hfill
 %  \subfloat[]{\includegraphics[width=0.46\textwidth]{ICCV/figures/CIFAR100_NMI.png}\label{figcifar100_b}}
%           \hfill
%   \subfloat[]{\includegraphics[width=0.46\textwidth]{ICCV/figures/CIFAR100_ARI.png}\label{figcifar100_c}}
%  \caption{Comparison of clustering performance on CIFAR-100 in terms of accuracy (a), NMI (b), and ARI (c).}
%\label{fig_cifar100}cifar100
%\end{figure}

%----------------------- CONCLUSION -----------------%
\section{Conclusion}  \label{conclusion}

In this paper, we present a new approach to jointly learn debiased representations and perform image clustering. Our framework is composed of two unsupervised neural networks: a deep statistical learning network and a deep clustering network. The network performs clustering and learns embedded features produced by deep neural networks. Updates in the network are done by optimizing a clustering loss with a statistical training distribution from the statistics network. To this end, we introduce a statistical pooling layer that learns to mitigate problems stemming from long-tailed data distribution by estimating the mean, variance, and cardinality of the feature space. Empirical experiments demonstrate the efficacy of the proposed method.

%\section{Acknowledgement}  

%This work has been funded in part by the German Federal Ministry of Education and Research (BMBF) under Grant No. 01IS18036A, Munich Center for Machine Learning (MCML).

\section{Acknowledgments}
ED is supported by the Helmholtz Association under the joint research school "Munich School for Data Science - MUDS" (Award Number HIDSS-0006).
M. R. and B. B. were supported by the Bavarian Ministry of Economic Affairs, Regional Development and Energy through the Center for Analytics – Data – Applications (ADA-Center) within the framework of BAYERN DIGITAL II (20-3410-2-9-8).
M. R. and B. B. were supported by the German Federal Ministry of Education and Research (BMBF) Munich Center for Machine Learning (MCML).

\bibliographystyle{IEEEtran}
\bibliography{ref}

\end{document}